\documentclass[journal]{IEEEtran}
\usepackage[table,xcdraw]{xcolor}
\usepackage{multirow}
\usepackage{xcolor,soul,framed} 

\colorlet{shadecolor}{yellow}
\usepackage[pdftex]{graphicx}
\graphicspath{{../pdf/}{../jpeg/}}
\DeclareGraphicsExtensions{.pdf,.jpeg,.png}

\usepackage[cmex10]{amsmath}
\usepackage{array}
\usepackage{mdwmath}
\usepackage{mdwtab}
\usepackage{eqparbox}
\usepackage{url}
\usepackage{hyperref}
\hypersetup{
            colorlinks=true,
            linkcolor=blue,
            anchorcolor=blue,
            citecolor=blue}
\usepackage[justification=centering]{caption}
\usepackage{float}
\usepackage{stfloats}
\usepackage{amsmath}

\usepackage{array}
\ifCLASSINFOpdf
\else
\fi
\begin{document}
%
\title{Training Robust Deep Physiological Measurement Models with Synthetic Video-based Data}
%
%
%

\author{

         Yuxuan~Ou, 
         Yuzhe~Zhang,
         Yuntang~Wang,
         Shwetak~Patel,
         Daniel~McDuf,
         Yuzhe Yang,
        Xin~Liu
 }       


%
%

\markboth{Journal of \LaTeX\ Class Files,~Vol.~14, No.~8, August~2015}%
{Shell \MakeLowercase{\textit{et al.}}: Bare Demo of IEEEtran.cls for IEEE Journals}
%



\maketitle

\begin{IEEEkeywords}
IEEE, IEEEtran, journal, \LaTeX, paper, template.
\end{IEEEkeywords}

\IEEEpeerreviewmaketitle
\begin{abstract}
Recent advances in supervised deep learning techniques have demonstrated the possibility to remotely measure human physiological vital signs (e.g., photoplethysmograph, heart rate) just from facial videos. However, the performance of these methods heavily relies on the availability and diversity of real labeled data.
Yet, collecting large-scale real-world data with high-quality labels is typically challenging and resource intensive, which also raises privacy concerns when storing personal bio-metric data.
Synthetic video-based datasets (e.g., SCAMPS~\cite{mcduff2022scamps}) with photo-realistic synthesized avatars are introduced to alleviate the issues while providing high-quality synthetic data.
However, there exists a significant gap between synthetic and real-world data, which hinders the generalization of neural models trained on these synthetic datasets. In this paper, we proposed several measures to add real-world noise to synthetic physiological signals and corresponding facial videos. We experimented with individual and combined augmentation methods and evaluated our framework on three public real-world datasets. Our results show that we were able to reduce the average MAE from 6.9 to 2.0.
\end{abstract}
\begin{figure*}[htbp]
\centering
\includegraphics[scale=0.7]{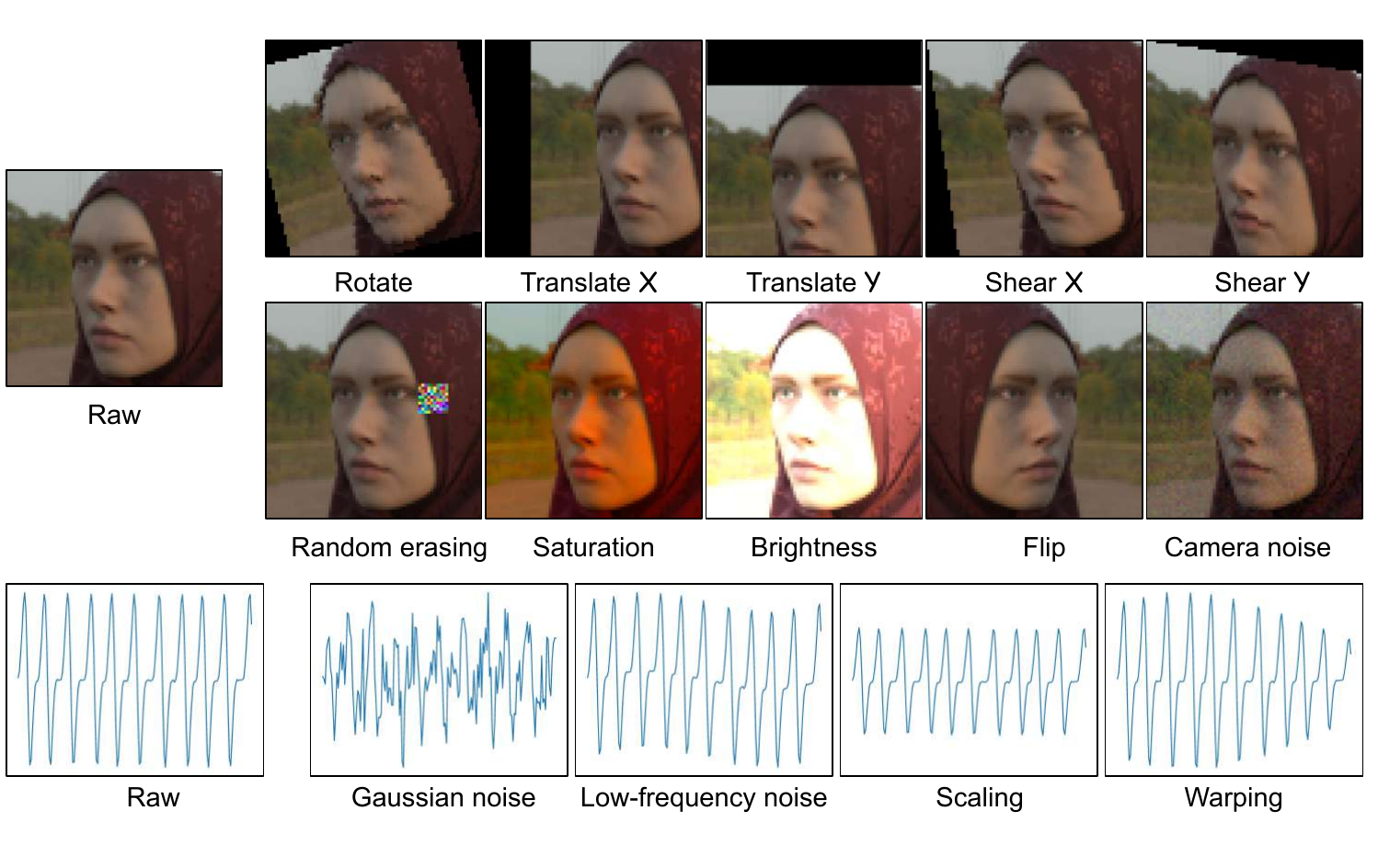}
\caption{\textbf{Visualization of video and signal operations.} In our study, we examine ten different video operations and four signal operations to simulate real-world noise and artifacts in digital sensor data. The video operations are divided into two categories: geometric transformations and appearance transformations. Geometric transformations include Rotate, Translate, Shear, and Flip, which mimic motion artifacts that can be present in video data. Appearance transformations include Random Erasing, Saturation, Brightness, and Camera Noise, which represent various types of noise that can arise from the digital camera and surrounding environment. In addition, we examine four signal operations: Gaussian noise, Low-frequency noise, Scaling, and Warping. Gaussian noise and Low-frequency noise are known to be present in real-world sensor data, while Scaling, and Warping are commonly used for time-series data augmentation} 
\end{figure*}

\section{Introduction}

Telehealth, or telemedicine, is a modern healthcare approach that utilizes technology to provide medical services remotely\cite{amjad2023review}. It brings numerous benefits to both patients and doctors. For doctors, telehealth eliminates the need for commuting between different inquiry sites, saving valuable time and allowing for more efficient use of resources. Patients, on the other hand, benefit from the convenience of remote consultations and the ability to have their health monitored from the comfort of their homes. Telehealth is revolutionizing healthcare delivery, enhancing accessibility and improving patient care\cite{sahoo2023distance}. 

Non-contact measurement of physiological signs is a core technology in telehealth that enables remote physicians to access and monitor vital physiological data without physical contact with the patient\cite{liu2023rppgmae}. Camera-based physiological monitoring has emerged as a promising approach, as well as various clinical and non-clinical applications, including neonatal monitoring, kidney dialysis, sleep monitoring, driver monitoring, and fake video detection\cite{kwon2023breathing}\cite{liu2023efficientphys}\cite{mcduff2023camera}. The technique offers the advantage of non-invasive and continuous monitoring of physiological signals, without requiring any physical contact with the subject. With the rapid advancements in computer vision and machine learning, several deep learning-based models have been proposed for remote monitoring of physiological signals using facial videos. These methods have shown great potential in measuring various physiological parameters, including heart rate, respiratory rate, and blood pressure, from video signals obtained using standard cameras. However, the performance of these models heavily relies on the availability and diversity of labeled data\cite{10.1145/3558518}.

 Supervised learning has emerged as the dominant approach for video-based physiological measurement, outperforming traditional signal processing methods. This is because supervised learning methods can learn complex temporal-spatial and colorspace representations\cite{10.1145/3558518}. Among supervised learning methods, convolutional neural networks are the most widely used and have achieved state-of-the-art performance in this task\cite{Spetlik2018VisualHR,Chaichulee,10.1007/978-3-030-01216-8_22,liu2020multi-task}. These networks learn representations by applying convolutions to the input frames. Unlike signal processing methods, supervised learning methods are more robust when dealing with sources of noise commonly encountered in real-world applications, such as head motions and ambient lighting changes.

Video-based physiological measurement using data-driven neural networks has shown promising results in predicting rPPG signals \cite{yang2023simper}. However, the accuracy and generalizability of these models heavily rely on the availability and diversity of large-scale and diverse datasets. Unfortunately, collecting such datasets is often time-consuming and presents privacy concerns. Real-world datasets also suffer from various limitations, including limited diversity in subject appearance, ambient illumination, and motions. To address these issues, researchers have introduced a synthetic dataset called SCAMPS, which contains 2,800 high-fidelity human simulations created by a graphics pipeline\cite{mcduff2022scamps}. While SCAMPS eases the burden of collecting data in various conditions and avoids privacy concerns, its performance still falls short when training on synthetic and testing on real-world datasets. The measurement results in the paper by McDuff et al. \cite{mcduff2022scamps} indicate that models trained on SCAMPS and tested on real-world datasets perform poorly compared to models trained on real-world datasets. This suggests that SCAMPS is not yet an adequate replacement for real-world datasets, and there is a significant room for improvement in terms of generalization from synthetic videos to real ones. Moreover, while SCAMPS offers the benefits of avoiding privacy concerns and the difficulties of collecting diverse real-world data, it is still limited in its ability to capture the full range of appearance and motion variations seen in real-world scenarios, which can affect model performance. Therefore, more efforts are needed to improve the quality and diversity of synthetic datasets for camera physiological measurement, to better support the development and deployment of reliable models in real-world applications.

The limitations of training solely on synthetic data are well-recognized, as there exists a significant gap between synthesized and real-world data, including videos and physiological signals \cite{mcduff2021synthetic}. As such, there is a critical need for methods that can effectively bridge the gap between simulated and real data, ensuring that models trained on synthetic data can accurately generalize to real-world scenarios. Addressing this gap is of utmost importance in ensuring the robustness and practicality of camera-based physiological measurement methods.

In this paper, we aim to address the simulation-to-real generalization gap in synthetic video-based physiological measurement datasets. Specifically, we investigate the effectiveness of adding real-world noise to computer graphics engine-generated synthesized datasets to improve the performance of deep learning models. By simulating real-world noise such as head movements, camera movements, and camera noises, we aim to improve the generalization ability of deep learning models trained on synthetic datasets. Our approach has the potential to reduce the reliance on costly and time-consuming real-world data collection, while also improving the accuracy and robustness of physiological measurements in real-world scenarios.

In summary, this paper makes four main contributions: First, we propose various approaches to adding real-world noise to synthetic physiological signals and corresponding facial videos. Second, we conduct a organized evaluation of various data augmentation techniques and summarize the most effective ones for video-based physiological measurement tasks on synthetic datasets. Third, we systematically study the effects of combining different data augmentation methods. Finally, we significantly improve heart rate prediction accuracy by training on the SCAMPS dataset, validating on SCAMPS, and testing on three most frequently used real-world datasets.

\section{Related Work}

\subsection{Synthetics data on other fields}

Synthetic data is a solution to several problems related to datasets used in machine learning tasks\cite{app11052158}. One problem is the difficulty in obtaining an adequate amount of data due to privacy restrictions\cite{Andriyanov_2020}. Another issue is that label generation is laborious work because raw datasets require manual processing\cite{app11052158}. Synthetic data are automatically generated by computer engines, saving researchers from manual labeling and allowing biases to be eliminated through presettings\cite{2021Synthetic}. Synthetic datasets are commonly used to simulate complex scenarios\cite{https://doi.org/10.48550/arxiv.2101.01394,2018Multi,2009Synthetic,https://doi.org/10.48550/arxiv.1911.05358} and overcome privacy issues\cite{https://doi.org/10.48550/arxiv.2208.09191,Cristina2017Temporal,https://doi.org/10.48550/arxiv.2104.02815}. Synthetic data has been used in many machine learning tasks\cite{DBLP:journals/corr/abs-2007-08781}, including human image or video analysis domains such as human detection\cite{ARANJUELO2021107105}, crowd counting\cite{wang2021pixel}, facial expression classification\cite{ahmed2019facial}, and smart phone user authentication\cite{buriro2021swipegan}. Several synthetic city scene datasets\cite{2016The,2016Virtual,2017Driving,2016Playing,2017Playing,https://doi.org/10.48550/arxiv.1807.06132} have been proposed to fuel the training of neural networks in tasks such as semantic instance segmentation, object detection, and tracking\cite{DBLP:journals/corr/abs-2007-08781}. Synthetic datasets also bring about controllability by generating data that considers real-world distributions of importance factors such as male-female ratio\cite{https://doi.org/10.48550/arxiv.2208.09191}. Finally, pure synthetic datasets can prevent hindrances caused by privacy concerns and information leaks\cite{feng2021gans,https://doi.org/10.48550/arxiv.2208.09191}.

\subsection{Synthetic data in physiological signal measurement}

In the field of physiological signal measurement, synthesized data has become increasingly prevalent in research, offering signal diversity and a priori knowledge of the signals. For instance, in \cite{6780577}, synthesized ballistocardiogram (BCG) signals were generated with physiological variability of heartbeat shape in mind, and were then used to verify a new method for detecting heart rate from BCG. Similarly, in \cite{9150925}, ECG and PPG data were simulated with a broad range of heart and respiration rates to test the performance of algorithms estimating these vital signs. In \cite{DONOSO20131628}, synthetic data provided a priori information about atrial activity content, helping to select the most representative source of this activity. Our synthetic dataset, SCAMPS, is unique in that it is the first dataset for vital signal measurement containing both synthesized videos and synthesized rPPG labels.

\subsection{General Data Augmentation}

Time series data augmentation techniques are widely used to improve the quantity and quality of training data in time series analysis tasks such as anomaly detection and time series classification, as deep neural networks rely heavily on large amounts of data, and many time series tasks suffer from inadequate labeled data. A taxonomy of time series augmentation methods was proposed in \cite{ijcai2021p631}.

In the time domain, Gaussian noise injection is the most commonly used method and has proven effective in many tasks. Window cropping involves segmenting the time series data into smaller windows and is commonly used in time series classification tasks with convolutional neural networks \cite{windowcrop}. Window warping randomly selects a time range and compresses or stretches it while keeping other time ranges unchanged. Flipping, which reverses the sign of the original label, is used in both classification and anomaly detection.

In the frequency domain, a recent work \cite{DBLP:journals/corr/abs-2002-09545} proposed adding perturbations to both the amplitude and phase spectra for data augmentation in time series anomaly detection using convolutional neural networks, and demonstrated performance improvements brought by this type of data augmentation.

Video augmentation methods have been widely used in various computer vision applications, including person re-identification and action recognition. In \cite{isobe2020intra}, Isobe et al. proposed an intra-clip data augmentation method for video-based person re-identification. The authors applied random cropping, flipping, and erasing to all the frames of a mini-batch clip, which helped improve the performance of person re-identification models by enhancing the diversity and robustness of the training data.
Another notable video augmentation method is VideoMix \cite{yun2020videomix}, proposed by Yun et al. VideoMix extends the CutMix method to video data augmentation by randomly selecting a patch from one video and replacing it with a patch from another video in the same mini-batch. This preserves the temporal consistency by keeping the patch size and position the same for all frames of each video clip. VideoMix has shown to be effective in improving the performance of various video-based tasks, such as action recognition and video classification. Other video augmentation methods include frame shuffling \cite{su2020unsupervised}, temporal jittering \cite{chu2019multi}, and motion blurring \cite{wang2020unsupervised}.

\subsection{Data Augmentation in physiological signal measurement}

Various noise types, including Gaussian, uniform, pink (low-frequency), and brown (high-frequency) noise, have been applied to rPPG labels in \cite{article} to prevent overfitting and enhance the generalization ability of deep learning models. Scaling and magnitude-warping, which multiply a coefficient to the magnitude of rPPG labels, are two other methods used for data augmentation \cite{article}.
For facial videos, randomly erasing some pixels from each frame is a data augmentation method used in the task of rPPG measurement \cite{9133501}. This technique reduces the effect of head movements on ROI occlusion.
In terms of data augmentation implemented on both videos and labels, \cite{2021Weakly} utilizes upsampling multi-scale spatial-temporal maps 1.5 or 2 times from neighboring videos and flipping both videos and labels in the time domain. This is used to increase the diversity of training samples and mitigate the difference between training and testing samples. Upsampling and downsampling both facial videos and rPPG labels are also mentioned in \cite{9133501}.

\section{Method}

This paper presents a comprehensive exploration of data augmentation techniques applied to labels (rPPG signals) and frames (video data) to enhance the performance of a remote photoplethysmography (rPPG) measurement model. Initially, the effects of individual operations are studied, followed by an evaluation of combinational operations, which include pairwise and a combination of eight carefully selected operations. These operations encompass Gaussian noise, warping, baseline wander, scaling, Poisson-Gaussian noise, shear, rotation, flipping, translation, random erasing, and brightness and saturation adjustments. The systematic investigation aims to optimize the model's accuracy and robustness in real-world settings.

In the initial phase of our study, we examine the impact of applying a single operation during the training process. The strength of the operation remains consistent within each batch. To facilitate a comprehensive understanding, we provide detailed descriptions of the fourteen selected operations, specifically tailored to suit our task, as outlined below.

In real-world datasets, the ground truth is acquired through sensor measurements. For instance, the UBFC-Phys dataset's labels, consisting of PPG waveforms, were obtained using a CMS50E transmissive pulse oximeter \cite{9346017}. These sensor signals may encompass various types of noise, such as baseline wander noise, muscle noise, and power line interference \cite{Rehman}. To simulate baseline wander noise, we add a sinusoidal wave to the label signal, with the magnitude of the wave ranging from 0 to 0.2 and its frequency varying from 0 to 0.5 Hz. Baseline wander is a low-frequency artifact commonly observed in PPG and ECG signal recordings, which can lead to signal modulations \cite{MEJIAMEJIA202269} \cite{gupta2015baseline}.

In our research, we introduce Additive Gaussian noise to the original signal to simulate its transmission over a channel with additive white Gaussian noise (AWGN). Gaussian noise with a mean of zero is added to the signal to replicate the noise encountered in the AWGN channel. For this work, we set the variance of the Gaussian noise to 0.5.

We explore two common data augmentation techniques for temporal signals: magnitude warping and scaling.

Magnitude warping involves multiplying each point in the signal by a random coefficient drawn from a Gaussian distribution. This operation preserves the temporal structure of the original signal but alters the amplitudes, resulting in a modified version. By applying this technique multiple times with different coefficients, we can create a dataset with numerous augmented samples, effectively increasing the dataset's size and diversity. For our research, we set the range of the standard deviation of the Gaussian distribution as (0, 0.25). Magnitude warping is widely used in various signal processing tasks, including speech recognition, image and video processing, and physiological signal analysis \cite{um2017data}.

On the other hand, scaling is a linear transformation applied to modify the amplitude of a signal by multiplying it with a scalar factor. This technique allows us to simulate variations in the intensity or strength of the signal. In our work, we set the range of the scaling factor as (0.75, 1.25). Scaling is another valuable data augmentation approach employed to enhance the performance of various signal processing applications.

The observed signal in digital imaging sensors can be represented as the sum of the original signal and a noise component, given by the equation \[ y = x + n \] To model the compound noise in CMOS/CCD imaging sensors, the Poisson-Gaussian noise model is commonly employed. This model consists of two parts: the Poissonian component, accounting for photon sensing, and the Gaussian component, representing stationary disturbances \cite{foi2008practical}. Typically, the Poissonian part is approximated as a Gaussian distribution with a signal-dependent variance \cite{jang2021c2n}. The noise component, denoted as $n$, follows a normal distribution with mean 0 and variance \[n \sim \mathcal{N}(0,\sigma_s^2x+\sigma_c^2)\]
where $\sigma_s^2$ and $\sigma_c^2$ represent the variances of the signal-dependent and signal-independent terms, respectively.

Spatial transformations, also known as geometric transformations, play a crucial role in addressing issues related to motion tolerance in rPPG measurement. These transformations encompass rotation, translation, shear, and flip, and their application enables us to preserve the value of the original signal in each pixel while altering the location of these pixels.

Rotation involves rotating points in the $xy$ plane either clockwise or counterclockwise through an angle $\theta$ with respect to the positive x-axis about the origin of a two-dimensional Cartesian coordinate system. On the other hand, translation shifts all points in the object in a straight line, maintaining a consistent direction of movement.
In the context of plane geometry, a shear mapping is a linear transformation that shifts each point proportionally to its signed distance from the line running parallel to the designated direction and passing through the origin. Each point undergoes a fixed shift in this specified direction.
Lastly, the horizontal flip operation involves flipping the frame horizontally from left to right. This transformation further enhances the robustness of rPPG measurement.
Table 1 illustrates the transformed coordinates resulting from the application of the aforementioned spatial transformations, showcasing their effects on the pixel positions in the image frame. By employing these transformations, we effectively mitigate motion-related challenges, thereby enhancing the accuracy and reliability of rPPG signal estimation in real-world scenarios.

\begin{table}[htbp]
	\centering
	\caption{\textbf{Coordinates of spatial transformations.} $(x,y)$ is the original coordinate.}  
	\begin{tabular}{cc}
		\hline\hline\noalign{\smallskip}	
		Spatial transformation & transformed coordinates\\
		\noalign{\smallskip}\hline\noalign{\smallskip}
		rotation & $(x\cos(\theta)-y\sin(\theta),x\sin(\theta)+y\cos(\theta))$  \\
        translate to x axis  & $(x+m, y)$ \\
        translate to y axis & $(x, y+m)$ \\
		shear to x axis  & $(x+my, y)$ \\
        shear to y axis  & $(x, mx+y)$ \\
        flip in x axis  & $(a-x,y) $ \\
		\noalign{\smallskip}\hline
	\end{tabular}
\end{table}

Random erasing is a widely employed data augmentation technique in image-related tasks, serving as a valuable tool for introducing variability into the dataset. This technique randomly selects a rectangular region within the image and replaces the original pixel values in that region with random values\cite{zhong2020random}. The beauty of this approach lies in its simplicity and parameter-free nature, making it easy to implement. For our specific task, we leverage the random erasing method to induce diversity by randomly erasing different regions in each frame. This strategic approach helps us avoid the risk of inadvertently erasing critical information consistently across all frames. As a result, our model benefits from a more comprehensive and varied dataset, contributing to improved performance and robustness in rPPG signal estimation. In our study, we set the area of the selected rectangular region to be 7*7, carefully determining an optimal size to strike a balance between introducing variability and preserving crucial features within the images. The incorporation of random erasing as part of our data augmentation strategy significantly enhances the accuracy and generalizability of our model, ensuring its suitability for real-world applications.

Random adjustment of brightness and saturation are widely employed data augmentation techniques in image and video processing tasks, offering effective means to increase dataset diversity and enhance model generalization. Brightness adjustment entails the addition or subtraction of a random value from each pixel in the image or video frame, mimicking changes in lighting conditions. On the other hand, saturation adjustment involves scaling the color saturation of the image or video frame by a random value, emulating variations in color vividness.
By applying these techniques during training, we introduce an array of lighting and color variations into the dataset, thereby enriching the training samples and enabling our model to better adapt to real-world scenarios. The dynamic nature of brightness and saturation adjustments ensures that our model learns to handle different lighting conditions and color intensities effectively, culminating in a more robust and accurate rPPG signal estimation.
In this study, we set the scale for brightness and saturation adjustments as (0.75, 1.25), thoughtfully determining a suitable range that strikes a balance between introducing diversity and avoiding excessive distortion. These carefully chosen scales contribute to the model's resilience in handling diverse visual conditions, making it better-equipped for real-world applications.

To comprehensively study the effects of these operations, we conduct a meticulous evaluation of combinational operations. Initially, we investigate the impact of operations combined pairwise. Drawing insights from the results of the single operation evaluations, we identify the most promising operations encompassing both video and rPPG signal operations. Subsequently, we rigorously evaluate the performance of every combination of two selected operations.
It is important to highlight that the order of applying these two operations is pivotal, except in specific scenarios: when one operation is a video operation and the other is a signal operation, or when both operations involve translation. Figure 2 provides a visual representation of this phenomenon, demonstrating how the processed images differ based on the order of transformations. For instance, applying rotation first and then translation yields distinct results compared to applying translation first and then rotation.
By meticulously studying the impact of various operation sequences, we gain crucial insights into the intricate interplay between video and rPPG signal operations, contributing to a more nuanced understanding of the overall performance of our model. Such insights equip us with valuable knowledge to design and optimize our model, enhancing its accuracy and robustness for real-world applications.

\begin{figure}[htbp]
\centering
\includegraphics[scale=0.6]{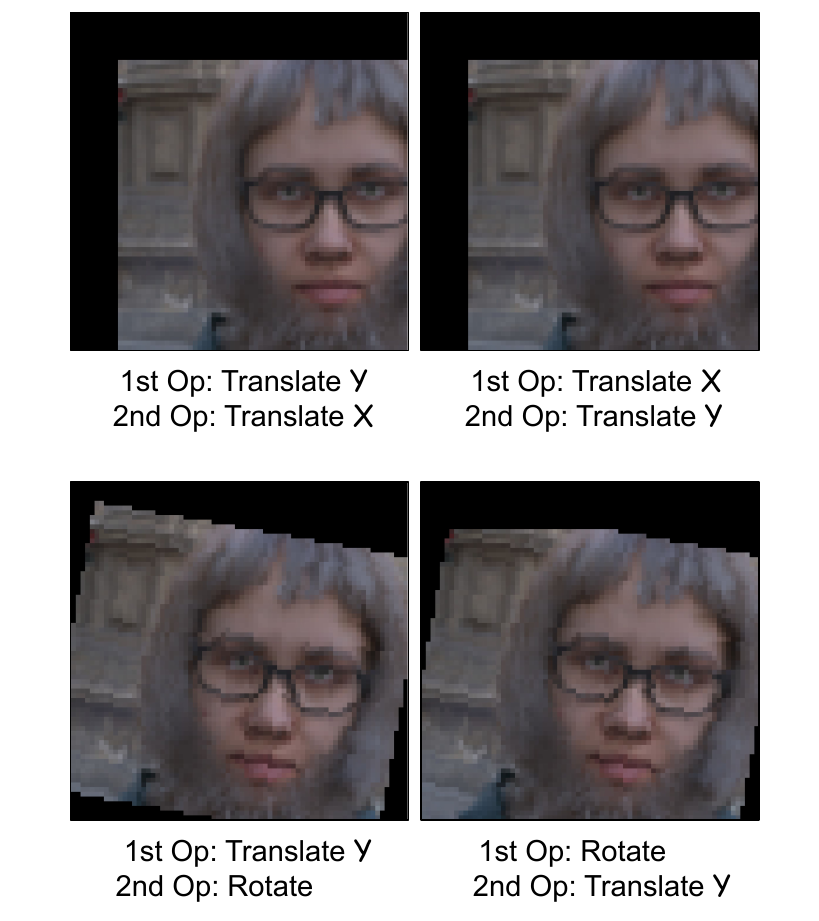}
\caption{\textbf{The order in which data augmentation operations are applied can have an impact on the resulting augmented data.} The images in the first row are identical, indicating that the order of Translate X and Translate Y operations does not affect the outcome. However, when the selected operations do not exclusively involve Translation, the sequence of applying these operations becomes significant. For instance, the resulting frames obtained by applying Translate and Rotate in different orders are distinct. This is also true when signal operations are involved in the augmentation process.} 
\end{figure}

\begin{figure*}[htbp]
\centering
\includegraphics[scale=0.7]{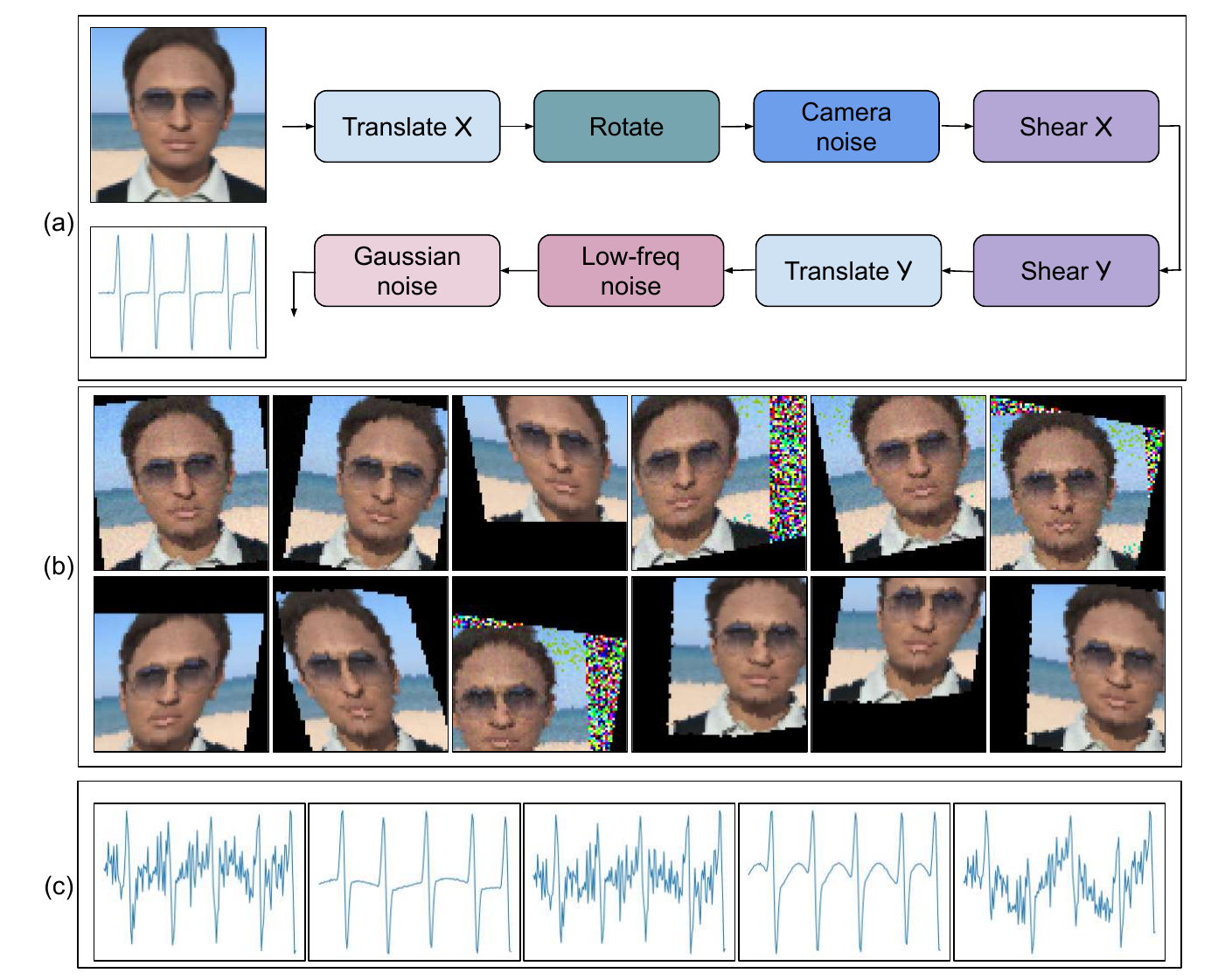}
\caption{\textbf{A flowchart of the proposed method.} Section (a) presents the input frame and label signal, followed by the sequential application of eight selected operations, listed with their names and order. The parameters for each operation are detailed in the appendix. Each operation is applied with a 0.5 probability. Section (b) displays a sample of resulting frames obtained from the operation sequence. In section (c), a set of resulting waveforms is shown. } 
\end{figure*}
Having thoroughly evaluated the performance of pairwise operations, we proceed to propose a combination of eight carefully selected operations. The sequence of these operations is determined based on the insightful findings obtained from the experiments conducted in pairwise combinations. To provide a clear visual representation of our proposed method, we present Figure 3, an illustrative depiction showcasing the sequential application of the eight operations. This comprehensive combination is thoughtfully designed to leverage the strengths of each operation and enhance the overall performance of our model for remote photoplethysmography (rPPG) measurement. By employing this strategically curated set of operations, we aim to achieve state-of-the-art results and advance the capabilities of rPPG signal estimation models in real-world scenarios.

\section{Experiment}
\subsection{Datasets}

\subsubsection{SCAMPS}
SCAMPS\cite{https://doi.org/10.48550/arxiv.2206.04197} comprises 2,800 videos, which corresponds to 1.68 million frames, and contains synchronized cardiac and respiratory signals. The facial pipeline used to generate the waveforms and videos is advanced and results in data that is highly realistic, almost resembling photographs. Within the videos, various factors that can affect the accuracy of the measurements are present, such as head movements, facial expressions, and changes in ambient lighting.

\subsubsection{PURE} The PURE dataset\cite{puredataset} comprises video recordings of six tasks performed by 10 participants (8 males, 2 females). The recordings were captured using an RGB eco274CVGE camera (SVS-Vistek GmbH) with a resolution of 640x480 and 60 Hz. The subjects were seated in front of the camera, illuminated with natural light from a window and positioned an average distance of 1.1 meters away from the camera. The recordings were collected under different motion conditions, and gold-standard measures of PPG and SpO2 were obtained with a pulse oximeter (CMS50E) attached to the finger.

\subsubsection{UBFC-RPPG} The UBFC-RPPG RGB video dataset\cite{ubfcdataset} was captured using a Logitech C920 HD Pro at 30Hz with a resolution of 640x480 in uncompressed 8-bit RGB format. To collect the gold-standard PPG data, a CMS50E transmissive pulse oximeter was used. The recordings were conducted indoors with a blend of sunlight and indoor illumination, and the subjects were seated one meter away from the camera.

\subsubsection{COHFACE} COHFACE \cite{cohfacedataset}is a dataset that was specifically created to facilitate research on physiological signal measurement from facial videos. It consists of more than 5,000 high-resolution (1080p) video clips of 100 participants (50 females and 50 males) performing various tasks, such as reading, counting, and recalling. The videos were captured with a Logitech C922 Pro Stream webcam at 30 frames per second. In addition to the video data, the dataset also provides synchronized gold-standard measurements of three physiological signals: PPG, ECG, and GSR. The PPG signal was measured using a non-contact method based on the remote photoplethysmography technique, which estimates the pulse signal from the changes in skin color caused by blood flow.  The dataset is notable for its diversity in terms of age, ethnicity, and skin tone, making it suitable for training and evaluating physiological signal measurement algorithms that are robust to individual differences.

\subsection{Baseline experiments}

To prepare the data for training, we process the raw frames of size 320x240 pixels by cropping the central 240x240 pixel region and downsampling them using a bilinear technique to obtain frames with a resolution of 72x72 pixels. Subsequently, we compute difference frames by applying a difference operation to subsequent frames. Both videos and signals are chunked into segments of length 180. Our supervised model, TSCAN, is trained using these frames with a learning rate of 0.001 and the ADAM optimizer for 10 training epochs on videos from the SCAMPS training set. The output of TSCAN is the predicted PPG waveform. To filter the output, we apply a 2nd-order Butterworth filter with cut-off frequencies of 0.75 and 2.5 Hz \cite{https://doi.org/10.48550/arxiv.2210.00716}. We apply Fast Fourier transform to the filtered signal to calculate the heart rate. To assess the model's generalization ability, we test it on three real-world datasets: UBFC-rPPG, PURE, and COHFACE. We use the SCAMPS validation set, and the heart rate estimation model with the lowest mean absolute error (MAE) on the validation set is chosen and evaluated on the test sets. The implementation was conducted in Python, and the experiments were performed on an NVIDIA GeForce RTX™ 3090 GPU with CUDA. The source code and experiment details can be found online.
\section{Result}
\begin{table}[H]
\caption{\textbf{Average results on three real-world datasets.} We highlight the best results in each column. "Single" indicates the application of Shear $X$ individually, and "Pairwise" indicates the application of camera noise followed by Shear $X$. The probability of applying these operations is 0.5. It can be observed that the performance of pairwise operations is better than single operations, and our proposed method shows the greatest improvement.}
\centering
\begin{tabular}{ccccc}
    \hline\hline\noalign{\smallskip}	
    Methods & MAE$\downarrow$ & RMSE$\downarrow$  & MAPE$\downarrow$& Pearson$\uparrow$  \\
    \noalign{\smallskip}\hline\noalign{\smallskip}
    Baseline & 6.904 & 15.118 &9.408 & 0.579  \\
    Single & 3.411 & 8.886 &4.241 & 0.831 \\
    Pairwise & 2.231 & 7.240 & 2.465 &0.901  \\
    Proposed & \textbf{2.008} & \textbf{6.102} & \textbf{2.284}&\textbf{0.936}  \\
    \noalign{\smallskip}\hline
\end{tabular}
\end{table}

The performance metrics for evaluating the results include mean absolute error (MAE), root mean squared error (RMSE), mean absolute percentage error (MAPE), and Pearson correlation, calculated between the predicted heart rate and gold standard measurements. Table 3 demonstrates that applying the proposed augmentation method, which involves sequentially applying eight operations, significantly improves the rPPG predicting performance. Specifically, MAE improved by 71 percent compared to the baseline result, from 6.904 to 2.008, RMSE improved by 60 percent, from 15.118 to 6.102, MAPE improved by 76 percent , from 9.408 to 2.284, and Pearson correlation improved by 62 percent, from 0.579 to 0.936. Table 3 also shows the most promising single operation and the combination of two operations. All four metrics show a performance improvement when more operations are combined, indicating that it is beneficial to compose operations.
Table 4 presents the result of applying one operation, and the best performance is achieved by shear in the x-axis, which reduces the MAE result from 6.904 to 3.411. Camera noise performs similarly to shear in the x-axis, with an MAE of 3.412. Except for camera noise, all of the operations that yield promising results are geometric transformations. Among label transformations, adding low-frequency noise outperforms other transformations in this category.

\begin{table}[H]
	\centering
	\caption{\textbf{MAE results of one augmentation operation and baseline.} Results are ordered by their average performance on three real-world datasets. For instance, the best individual operation is Shear to $X$ axis and the worst operation is Flip.}
	  
	\begin{tabular}{ccccc}
		\hline\hline\noalign{\smallskip}	
		 \quad& UBFC  & PURE  & COHFACE  & Average \\
		\noalign{\smallskip}\hline\noalign{\smallskip}
		Shear $X$  &2.050 & 2.808  & 5.375 & 3.411 \\
  Camera noise&2.825 & 4.409 &3.002 & 3.412\\
    Translate $X$&2.092 & 3.470 &4.995 &3.519 \\
     Rotate & 2.615& 3.605 &4.702 & 3.640\\
       Translate $Y$ &2.322 & 4.409 &6.093 &4.275 \\
         Shear $Y$ &2.908 &4.014  &6.364 &4.429 \\
           Random erasing &2.929 &3.307  &10.166 &5.467 \\
            Low-freq noise  &5.608 &3.999  &9.034 & 6.214\\
               Brightness &5.022 &5.176  &8.598 &6.265 \\
                 Gaussian noise &1.904 & 4.640  &13.751 &6.765 \\
                Saturation     &4.415 & 4.580 &11.649 & 6.881\\
                Baseline& 4.268 & 4.454 & 11.989&6.904 \\
                Warping       &2.783 &4.342  &14.956 &7.360 \\
                   Scaling      &3.201 &4.752  &15.128 &7.693 \\
                         Flip  &6.842 &5.139  &12.158 &8.046 \\

  \noalign{\smallskip}\hline
	\end{tabular}

\end{table}

\begin{table}[H]
	\centering
	\caption{\textbf{Baseline results on three real-world datasets and the average results.} The baseline results indicate a significant sim-real gap between synthetic and real-world data, highlighting the need for methods to bridge this sim-real divide.}
	 
	\begin{tabular}{ccccc}
		\hline\hline\noalign{\smallskip}	
		Dataset & MAE$\downarrow$ & RMSE$\downarrow$  & MAPE$\downarrow$& Pearson $\uparrow$ \\
		\noalign{\smallskip}\hline\noalign{\smallskip}
		UBFC & 4.268 & 12.041 &4.232 & 0.816  \\
        PURE & 4.454 & 14.821 &5.441 & 0.783 \\
		COHFACE & 11.989 & 18.492&18.552&0.137 \\
        Average & 6.904 & 15.118 &9.408 & 0.579\\
		\noalign{\smallskip}\hline
	\end{tabular}
\end{table}

\begin{figure*}[b]
\centering
\includegraphics[scale=0.73]{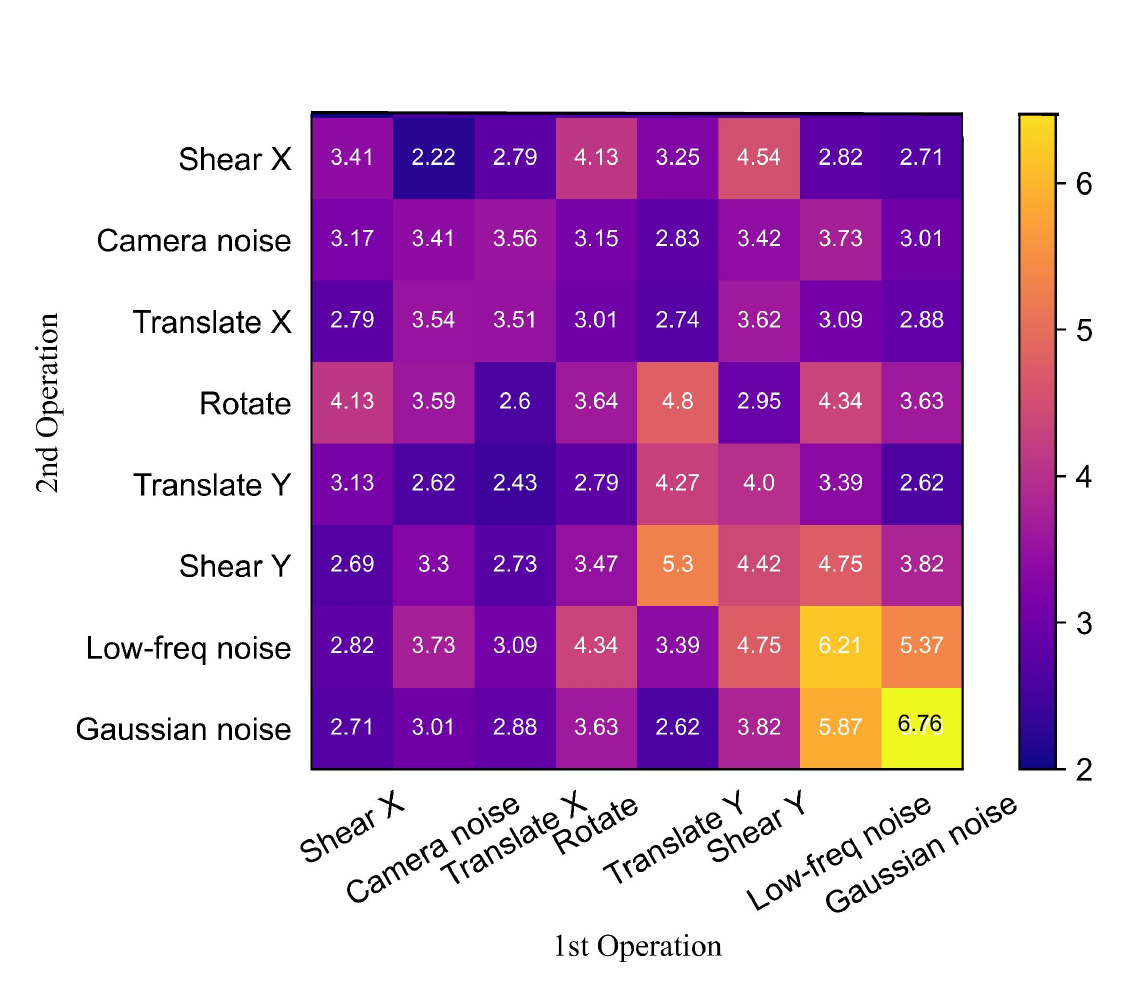}
\caption{\textbf{ Average MAE results on three real-world datasets with operations applied individually or in pairs.} The heatmap represents the MAE values with corresponding colors, where lighter colors correspond to larger MAE values, indicating poor performance, and darker colors represent smaller MAE values, indicating better performance. Notably, colors on the diagonal line from the upper-left to lower-right corner exhibit the lightest colors in their respective row or column, highlighting the importance of combining operations for achieving improved performance.} 
\end{figure*}
To comprehend the effects of various data augmentations and the importance of their combinations, we examine the performance when each augmentation is applied in pairs. Heatmap in figure 4 shows the result of individual and composed augmentations. The colors of blocks located on the diagonal line from the upper-left to lower-right corner are the lightest or second lightest in their respective row and column. This shows that composed augmentations outperform individuals. The augmentation scheme that yielded the greatest improvement is the composition of shear in the x-axis and camera noise, the two individuals with the greatest improvements. We conclude that pairwise is better than individual augmentations. To promote our finding, we propose adopting a combination of eight selected augmentations, which includes the most promising six video operations and two label operations. The sequence of the eight operations is determined by the results in figure 4. Table 5 displays the result of this method, which yields the best predicting result, with a 10 percent improvement compared to the best pairwise augmentation scheme.

\begin{table}[H]
	\centering
	\caption{\textbf{Results of the proposed method and improvement compared to baseline.} Our proposed method has demonstrated a significant improvement in performance, resulting in increased generalization ability of the model.}
	  
	\begin{tabular}{ccccc}
		\hline\hline\noalign{\smallskip}	
		Dataset & MAE$\downarrow$ & RMSE$\downarrow$  & MAPE$\downarrow$& Pearson$\uparrow$  \\
		\noalign{\smallskip}\hline\noalign{\smallskip}
		UBFC & 1.799 & 3.723 &1.928 & 0.980  \\
        PURE & 2.256 & 9.435 &2.380 & 0.919 \\
		COHFACE & 1.970 & 5.148&2.551&0.909 \\
        Average  & 2.008 & 6.102 & 2.284&0.936  \\
        Improvement & $\downarrow$4.896 &$\downarrow$9.016&$\downarrow$7.127&$\uparrow$0.357\\
		\noalign{\smallskip}\hline
	\end{tabular}
\end{table}

\section{Discussion}
\begin{figure*}[htbp]
\centering
\includegraphics[scale=0.55]{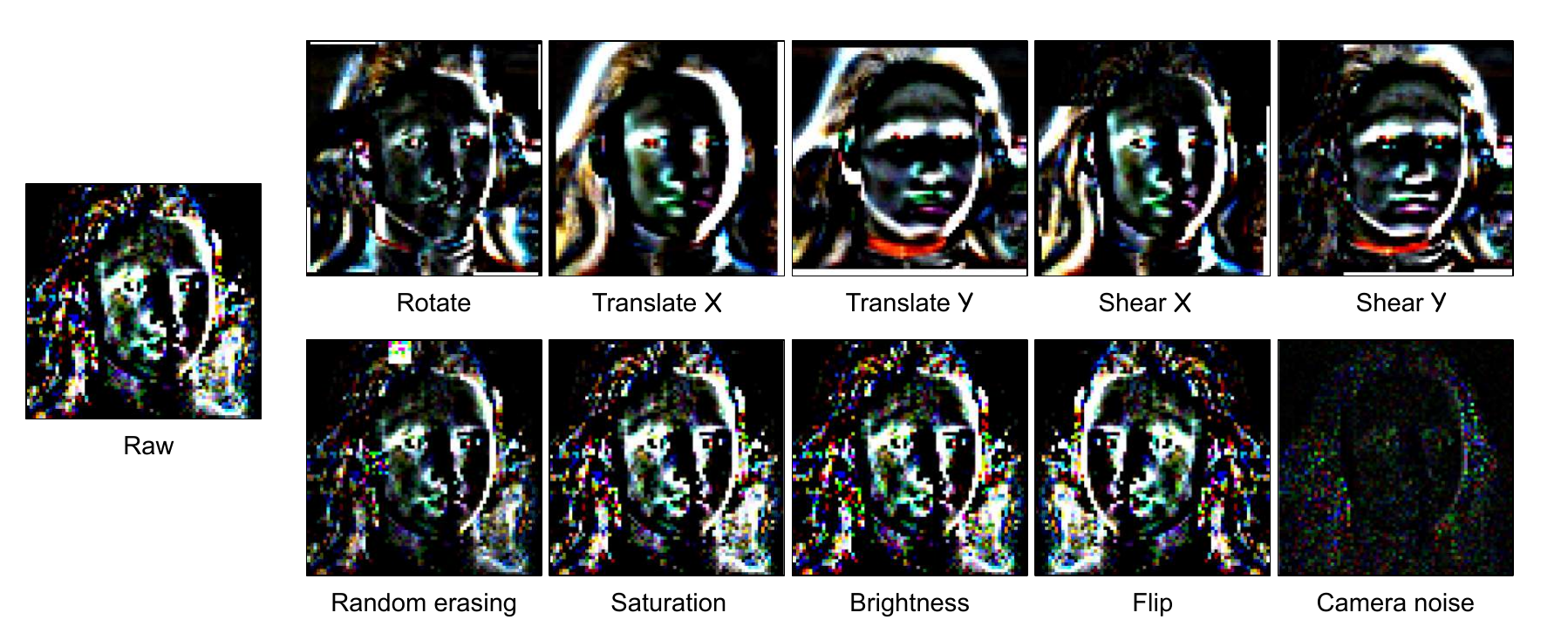}
\caption{\textbf{Visualization of difference frames.} Geometric transformations exhibit a significant difference when compared to the raw frame, whereas appearance transformations do not show a substantial deviation from the raw frame.} 
\end{figure*}

In this study, we first evaluated the performance of different data augmentation techniques on both synthetic videos and rPPG signals. 

As for video operations, Our results showed that flipping from left to right decreases performance, while geometric transformations and camera noise bring about significant improvements, and appearance transformations bring about less improvement. This may be because the input is a difference frame, and applying geometric transformations results in a more diverse set of difference frames, which can help improve the robustness of the model. 
 Flipping the input data from left to right resulted in worse results. The issue may be related to differences in how the human body is structured on the left and right sides. Heart rate can differ between the left and right sides of the body because the heart is slightly asymmetrical, and the left side is responsible for pumping oxygenated blood to the body. In contrast, the right side pumps deoxygenated blood to the lungs. This could impact how the model captures changes in blood flow during the rPPG measurement.

For rPPG signal operations, adding Gaussian noise and baseline wander effectively improved the models' accuracy. However, scaling and warping were ineffective in improving the performance of rPPG prediction models.
Adding Gaussian noise and baseline wander can work with synthetic rPPG signals because they introduce variations similar to what would be expected in real-world scenarios. Gaussian noise is a type of random noise that can be present in physiological signals due to external factors or internal physiological processes. On the other hand, baseline wander can be caused by various factors, such as breathing or movement, resulting in a slow drift of the signal baseline. Therefore, adding these variations can make the synthetic signals more realistic and improve the performance of machine learning models trained on them. Additionally, since these augmentations do not introduce any drastic changes to the signal shape or frequency content, they are less likely to cause the model to overfit the augmented data.
Although the addition of Gaussian noise and baseline wander improved rPPG prediction models, further research is needed to determine the optimal levels of noise and wander that can be added to the signals to maximize model performance without introducing too much distortion. 

Our findings suggest that the ineffectiveness of scaling could be due to a lack of real-world variability and different noise characteristics. The synthetic dataset is generated based on simplified assumptions or models, which may not fully capture the variability and complexity of real-world data. Therefore, applying scaling on synthetic data may generalize poorly to real-world data. Synthetic datasets may have different noise characteristics than real-world data, which can affect the effectiveness of scaling. 

Warping is a common data augmentation technique for improving machine learning models' performance. However, our experiments showed that warping did not work well for our synthetic dataset. This may be because the signals after warping have a different noise distribution than the real-world data. The synthetic dataset was generated with a specific noise characteristic, which differed from the real-world data, making warping ineffective.

Our results suggest that applying video and signal data augmentation can effectively improve the performance of rPPG signal estimation models. However, our findings also highlight the importance of carefully selecting the types of augmentations to apply. Specifically, we recommend focusing on geometric transformations and operations that simulate real-world noise, which can significantly improve performance. Additionally, our results suggest that some operations work in real-world datasets, they may result poorly in synthetic datasets.

While individual operations improved the algorithm's performance, the combination of operations had a more significant impact on the accuracy of the results.
In the heatmap, the values at the diagonal line are the largest or second largest in its row and column, indicating that pairwise operation can improve the algorithm's performance better. 

The proposed method, which is the combination of 8 selected operations, has significantly improved performance compared to using operations individually or in pairs. This result suggests that the selected operations complement each other and form a powerful method for the task. The combination of operations allows for a more diverse and robust feature representation, capturing different aspects of the data and making the model more adaptable to variations in the input. Furthermore, the combination of operations can help address each operation's limitations, such as overfitting or lack of generalization.

Our findings suggest that data augmentation can be a valuable tool for improving the accuracy and robustness of rPPG signal estimation models.
The results demonstrate the importance of combining multiple operations to achieve state-of-the-art performance. The proposed method can be used as a baseline for future research and can be further extended by incorporating additional operations or optimizing the combination of existing ones.
\section{Conclusion}

In this paper, we present a systematic evaluation of various data augmentation techniques to enhance the performance of rPPG signal estimation models in real-world settings. The study specifically focuses on the augmentation of both synthetic videos and rPPG signals. By examining the effectiveness of different augmentation approaches, we aim to provide insights into the most effective strategies for improving the accuracy and reliability of rPPG signal estimation.
The experimental results obtained from our evaluation highlight the substantial impact of applying geometric transformations and simulating real-world noise. Specifically, techniques such as camera noise and baseline wander simulation exhibit significant potential for enhancing the accuracy of the models. Moreover, we find that the combination of multiple augmentation operations yields even more promising outcomes, demonstrating the importance of leveraging a diverse set of augmentation techniques.

This study makes a significant contribution by showcasing the effectiveness of synthetic data in training rPPG signal prediction models. Synthetic data offers a cost-effective and scalable approach for generating large datasets to train machine learning models. By leveraging synthetic data, we not only enhance the size of the training dataset but also augment real-world datasets, thereby improving the models' resilience to variations in input data. To bridge the sim-real gap, we introduce real-world noises into synthetic videos and physiological signals, ensuring a more realistic representation of the data. Through the integration of eight combinational operations, our proposed method empowers models trained and validated on synthetic data to achieve state-of-the-art performance, comparable to models trained on real-world datasets.

The current approach presented in this study serves as a promising foundation for future investigations in the field. 
 Additional techniques can be incorporated or the existing combination can be refined to enhance the model's performance even further. It is essential to strike a balance when introducing noise into the data, as finding the optimal amount is crucial. Striving for an appropriate level of noise allows for improved model performance without introducing excessive distortion. This aspect warrants further research and consideration in future studies.


%


\section{}



\ifCLASSOPTIONcaptionsoff
  \newpage
\fi



%

\bibliographystyle{IEEEtran}
\bibliography{IEEEabrv, Bibliography}




%




\begin{IEEEbiographynophoto}{}
\end{IEEEbiographynophoto}





\end{document}